\documentclass{article}

\PassOptionsToPackage{numbers}{natbib}
\usepackage[preprint]{neurips_2022}

\usepackage[utf8]{inputenc} % allow utf-8 input
\usepackage[T1]{fontenc}    % use 8-bit T1 fonts
\usepackage{hyperref}       % hyperlinks
\usepackage{url}            % simple URL typesetting
\usepackage{booktabs}       % professional-quality tables
\usepackage{amsfonts}       % blackboard math symbols
\usepackage{nicefrac}       % compact symbols for 1/2, etc.
\usepackage{microtype}      % microtypography
\usepackage{xcolor}         % colors

\usepackage{mathrsfs}
\usepackage{amsmath}
\usepackage{graphicx}
\usepackage{multirow}
\usepackage{amssymb}

\usepackage[symbol]{footmisc} % for control footnote markers
\usepackage{float}  % for figures in Appendix

\title{Understanding Masked Image Modeling via Learning Occlusion Invariant Feature}

\author{Xiangwen Kong \\
        MEGVII Technology, \\
        Beijing, China \\
        \texttt{kongxiangwen@megvii.com}
        \And
        Xiangyu Zhang\textsuperscript{$\dagger$} \\
        MEGVII Technology, \\
        Beijing, China \\
        \texttt{zhangxiangyu@megvii.com}
}

\begin{document}

\renewcommand{\thefootnote}{\fnsymbol{footnote}}
\footnotetext[2]{Corresponding author. This work is supported by The National Key Research and Development Program of China (No. 2017YFA0700800) and Beijing Academy of Artificial Intelligence (BAAI).}
\renewcommand*{\thefootnote}{\arabic{footnote}}

\maketitle

\begin{abstract}

Recently, Masked Image Modeling (MIM) achieves great success in self-supervised visual recognition. However, as a reconstruction-based framework, it is still an open question to understand how MIM works, since MIM appears very different from previous well-studied siamese approaches such as contrastive learning. In this paper, we propose a new viewpoint: MIM implicitly learns occlusion-invariant features, which is analogous to other siamese methods while the latter learns other invariance. By relaxing MIM formulation into an equivalent siamese form, MIM methods can be interpreted in a unified framework with conventional methods, among which only a) data transformations, i.e. what invariance to learn, and b) similarity measurements are different. Furthermore, taking MAE (He \emph{et al.}~\cite{mae}) as a representative example of MIM, we empirically find the success of MIM models relates a little to the choice of similarity functions, but the learned occlusion invariant feature introduced by masked image -- it turns out to be a favored initialization for vision transformers, even though the learned feature could be less semantic. We hope our findings could inspire researchers to develop more powerful self-supervised methods in computer vision community. 

\end{abstract}

\section{Introduction}
\label{sec:intro}

\emph{Invariance} matters in science \cite{kosmann2011noether}. In self-supervised learning, invariance is particularly important: since ground truth labels are not provided, one could expect the favored learned feature to be invariant (or more generally, equivariant \cite{dangovski2021equivariant}) to a certain group of transformations on the inputs. Recent years, in visual recognition one of the most successful self-supervised frameworks -- \emph{contrastive learning} \cite{cpc,cmc,dosovitskiy2014discriminative} -- benefits a lot from \emph{learning invariance}. The key insight of contrastive learning is, because recognition results are typically insensitive to the deformations (e.g. cropping, resizing, color jittering) on the input images, a good feature should also be invariant to the transformations. Therefore, contrastive learning suggests minimizing the distance between two (or more \cite{dino}) feature maps from the augmented copies of the same data, which is formulated as follows:
\begin{equation}
     \mathop{\mathrm{min}}_\theta \mathop{\mathbb{E}}_{x\sim \mathcal{D}} \mathcal{M} \left(z_1, z_2 \right), \quad z_1=f_\theta(\mathcal{T}_1(x)), \quad z_2=f_\theta (\mathcal{T}_2(x)),
\label{eq:contrastive}
\end{equation}
where $\mathcal{D}$ is the data distribution; $f_\theta(\cdot)$ means the \emph{encoder network} parameterized by $\theta$; $\mathcal{T}_1(\cdot)$ and $\mathcal{T}_2(\cdot)$ are two transformations on the input data, which defines what invariance to learn; $\mathcal{M}(\cdot,\cdot)$ is the \emph{distance function}\footnote{Following the viewpoint in \cite{simsiam}, we suppose distance functions could contain parameters which are jointly optimized with Eq.~\ref{eq:contrastive}. For example, weights in \emph{project head} \cite{simclr} or \emph{predict head} \cite{byol,simsiam} are regarded as a part of distance function $\mathcal{M}(\cdot)$.  } 
(or \emph{similarity measurement}) to measure the similarity between two feature maps $z_1$ and $z_2$. Clearly, the choices of $\mathcal{T}$ and $\mathcal{M}$ are essential in contrastive learning algorithms. Researchers have come up with a variety of alternatives. For example, for the transformation $\mathcal{T}$, popular methods include random cropping \cite{bachman2019learning,moco,simclr,byol}, color jittering \cite{simclr}, rotation \cite{metzger2020evaluating,gidaris2018unsupervised}, jigsaw puzzle \cite{jigsaw}, colorization \cite{zhang2016colorful} and etc. For the similarity measurement $\mathcal{M}$, \emph{InfoMax principle} \cite{bachman2019learning} (which can be implemented with \emph{MINE} \cite{mine} or \emph{InfoNCE loss} \cite{cpc,moco,simclr,mocov2}), feature de-correlation \cite{barlowtwins,vicreg}, asymmetric teacher \cite{byol,simsiam}, \emph{triplet loss} \cite{li2021triplet} and \emph{etc.}, are proposed. 

Apart from contrastive learning, very recently \emph{Masked Image Modeling} (\emph{MIM}, \emph{e.g.} \cite{beit}) quickly becomes a new trend in visual self-supervised learning. Inspired by \emph{Masked Language Modeling} (\cite{bert}) in \emph{Natural Language Processing}, MIM learns feature via a form of \emph{denoising autoencoder} \cite{vincent2008extracting}: images which are occluded with random \emph{patch masks} are fed into the encoder, then the decoder predicts the original embeddings of the masked patches: 
\begin{equation}
\mathop{\mathrm{min}}_{\theta, \phi} \mathop{\mathbb{E}}_{x\sim \mathcal{D}} \mathcal{M} \left( d_\phi(z), x \odot (1-M) \right), \quad z = f_\theta(x \odot M), 
\label{eq:mim}
\end{equation}
where ``$\odot$'' means element-wise product; $M$ is \emph{patch mask}
\footnote{So ``$x\odot M$'' represents ``unmasked patches'' and vice versa.}; $f_\theta(\cdot)$ and $d_\phi(\cdot)$ are \emph{encoder} and \emph{decoder} respectively; $z$ is the learned representation; $\mathcal{M}(\cdot,\cdot)$ is the \emph{similarity measurement}, which varies in different works, \emph{e.g.} $l2$-distance \cite{mae}, \emph{cross-entropy} \cite{beit} or \emph{perceptual loss} \cite{dong2021peco} in \emph{codebook space}. Compared with conventional contrastive methods, MIM requires fewer effort on tuning the augmentations, furthermore, achieves outstanding performances especially in combination with \emph{vision transformers} \cite{dosovitskiy2020image}, which is also demonstrated to be scalable into large vision models \cite{mae,li2022exploring}. 

In this paper, we aim to build up a \emph{unified} understanding framework for \emph{MIM} and \emph{contrastive learning}. Our motivation is, even though MIM obtains great success, it is still an open question how it works. 
Several works try to interpret MIM from different views, for example, \cite{mae} suggests MIM model learns "rich hidden representation" via reconstruction from masked images; afterwards, \cite{understand_mae} gives a mathematical understanding for \emph{MAE} \cite{mae}. However, what the model learns is still not obvious. 
The difficulty lies in that MIM is essentially \emph{reconstructive} (Eq.~\ref{eq:mim}), hence the supervision on the learned feature ($z$) is \emph{implicit}. In contrast, contrastive learning acts as a \emph{siamese} nature (Eq.~\ref{eq:contrastive}), which involves \emph{explicit} supervision on the representation. If we manage to formulate MIM into an equivalent siamese form like Eq.~\ref{eq:contrastive}, MIM can be \emph{explicitly} interpreted as learning a certain \emph{invariance} according to some \emph{distance measurement}. We hope the framework may inspire more powerful self-supervised methods in the community. 

In the next sections, we introduce our methodology. Notice that we do not aim to set up a new state-of-the-art MIM method, but to improve the understanding of MIM frameworks. Our findings are concluded as follows:
\begin{itemize}
\item We propose \emph{RelaxMIM}, a new \emph{siamese} framework to approximate the original \emph{reconstructive} MIM method. In the view of RelaxMIM, MIM can be interpreted as a special case of contrastive learning: the data \emph{transformation} is random patch masking and the \emph{similarity measurement} relates to the decoder. In other words, \textbf{MIM models intrinsically learn occlusion invariant features}.
\item Based on RelaxMIM, we replace the similarity measurement with simpler \emph{InfoNCE loss}. Surprisingly, the performance maintains the same as the original model. It suggests that the reconstructive decoder in MIM framework does not matter much; other measurements could also work fine. Instead,  \textbf{patch masking may be the key to success}. 
\item To understand why patch masking is important, we perform MIM pretraining on very few images (\emph{e.g.} only \textbf{1} image), then finetune the encoder with supervised training on full ImageNet. Though the learned representations lack of semantic information after pretraining, the finetuned model still significantly outperforms those training from scratch. We hypothesize that the encoder learns \emph{data-agnostic} occlusion invariant features during pretraining, which could be a favored initialization for finetuning. 
 \end{itemize}

\section{MIM intrinsically learns occlusion invariant feature}
\label{sec:relax}

In this section, we mainly introduce how to approximate \emph{MIM} formulation (Eq.~\ref{eq:mim}) with a siamese model. For simplicity, we take \emph{MAE} \cite{mae} as an representative example of MIM, in which the \emph{similarity measurement} is simply $l2-$distance on the \emph{masked} patches. Other MIM methods can be analyzed in a similar way. Following the notations in Eq.~\ref{eq:mim}, the loss function for MAE training is\footnote{In original MAE \cite{mae}, the encoder network only generates tokens of unmasked patches and the decoder only predict the masked patches during training. In our formulations, for simplicity we suppose both networks predict the \emph{whole} feature map; we equivalently extract the desired part via proper \emph{masking} if necessary. }:
\begin{equation}
    L(x, M) = \| d_\phi(f_\theta(x \odot M)) \odot (1-M)  - x \odot (1-M) \|^2.
\label{eq:mae}
\end{equation}

\begin{figure}
  \centering
  \includegraphics[width=1.0\textwidth]{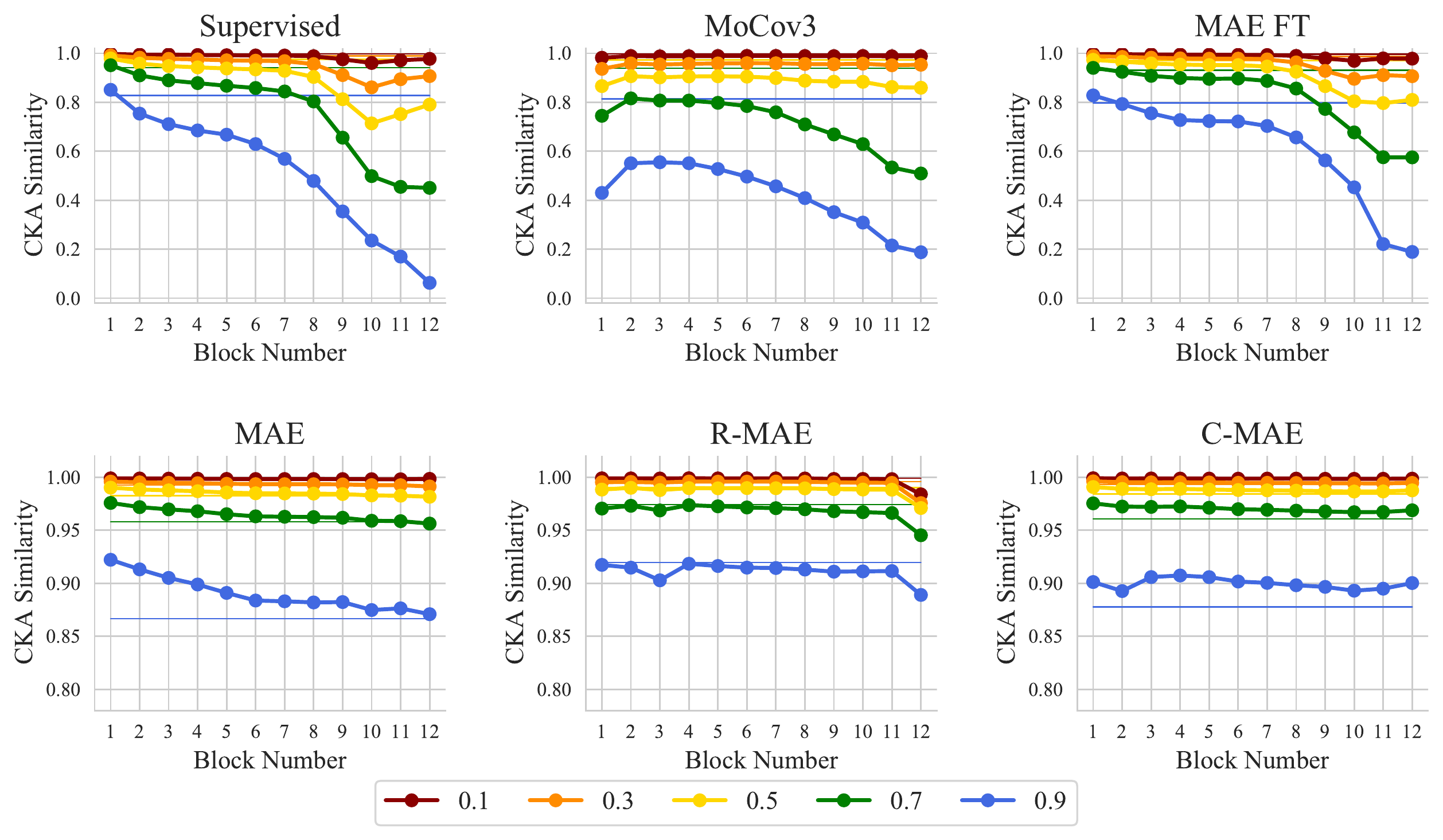}
  \caption{CKA similarity between the representations generated by the masked image and the full image respectively under different mask ratios. \emph{FT} means finetuned model on ImageNet; and other models except ``Supervised'' are self-supervised pretrained models. }
  \label{fig:cka}
\end{figure}

Let us focus on the second term. Typically, the dimension of feature embedding is much larger than dimension of input image, thus the encoder (at least) has a chance to be lossless \cite{li2022exploring}. That means for the encoder function $f_\theta(\cdot)$, there exists a network $d'_{\phi'}(\cdot)$ parameterized by $\phi'$ that satisfying $d'_{\phi'}(f_\theta(x \odot (1-M))) \odot (1-M) \approx x \odot (1-M)$. Then, we rewrite Eq.~\ref{eq:mae} in the following equivalent form:
\begin{equation}
\begin{split}
  L(x, M) &= \| d_\phi(f_\theta(x \odot M)) \odot (1-M) - d'_{\phi'}(f_\theta(x \odot (1-M))) \odot (1-M) \|^2 \\
  s.t. \quad \phi' &= \mathop{\arg\min}\limits_{\phi'} \mathop{\mathbb{E}}_{x'\sim \mathcal{D}} \|d'_{\phi'}(f_\theta(x' \odot (1-M))) \odot (1-M) - x' \odot (1-M) \|^2 
\end{split}
\label{eq:twostep_mae}
\end{equation}

Eq.~\ref{eq:twostep_mae} can be further simplified. Notice that $d'_{\phi'}(\cdot)$ just approximates the ``inverse'' (if exists) of $f_\theta(\cdot)$, there is no reason to use a different architecture from $d_{\phi}(\cdot)$. So we let $d' = d$. Then we define a new \emph{similarity measurement}:
\begin{equation}
    \overline{\mathcal{M}_{\phi, \phi'}}(z_1, z_2) \triangleq \| (d_\phi (z_1) - d_{\phi'} (z_2) ) \odot (1-M) \|^2,
\label{eq:new_measurement}
\end{equation}
and transformations: 
\begin{equation}
    \mathcal{T}_1(x) = x \odot M, \quad \mathcal{T}_2(x) = x \odot (1 - M),
\label{eq:new_transformations}
\end{equation}
hence Eq.~\ref{eq:twostep_mae} equals to:
\begin{equation}
\begin{split}
L(x, M; \theta, \phi) &= \overline{\mathcal{M}_{\phi, \phi'}}(f_\theta(\mathcal{T}_1(x)), f_\theta(\mathcal{T}_2(x))) \\
s.t. \quad \phi' &= \mathop{\arg\min}\limits_{\phi'} \mathop{\mathbb{E}}_{x'\sim \mathcal{D}} \|(d_{\phi'}(f_\theta(\mathcal{T}_2(x'))) - \mathcal{T}_2(x') ) \odot (1-M) \|^2.
\end{split}
\label{eq:disc_mae}
\end{equation}
We name Eq.~\ref{eq:disc_mae} \textbf{siamese form of MAE}. 

\paragraph{Discussion.} Eq.~\ref{eq:disc_mae} helps us to understand MIM from a \emph{explicit view}. Compared Eq.~\ref{eq:disc_mae} with Eq.~\ref{eq:contrastive}, the formulation can be viewed as a special case of \emph{contrastive learning}: the loss aims to minimize the differences between the representations derived from two masking \emph{transformations}. Therefore, we conclude that \textbf{MIM pretraining encourage occlusion invariant features}. The decoder joints as a part of the \emph{similarity measurement} (see Eq.~\ref{eq:new_measurement}), which is reasonable: since it is difficult to define a proper distance function directly in the latent space, a feasible solution is to project the representation back into the image space, because similarities like $l2$-distance in image space are usually explainable (analogous to \emph{PSNR}). In addition, the constraint term in Eq.~\ref{eq:disc_mae} can be viewed as standard \emph{AutoEncoder} defined on the space of $\mathcal{T}_2(x)$, which guarantees the projection $d_{\phi'}(\cdot)$ to be informative, avoiding collapse of the similarity measurement.

Although Eq.~\ref{eq:disc_mae} \emph{explicitly} uncovers the invariant properties of MIM in theory, it is a drawback that Eq.~\ref{eq:disc_mae} involves a nested optimization, which is difficult to compute. We thus propose a \emph{relaxed form} of Eq.~\ref{eq:disc_mae}, named \emph{R-MAE} (or \emph{RelaxMIM} in general):
\begin{equation}
    \mathop{\mathrm{min}}_{\theta, \phi, \phi'} \mathop{\mathbb{E}}_{x \sim \mathcal{D}} \\
    \overline{\mathcal{M}_{\phi, \phi'}}(f_\theta(\mathcal{T}_1(x)), f_\theta(\mathcal{T}_2(x))) + \\
    \lambda \|(d_{\phi'}(f_\theta(\mathcal{T}_2(x))) - \mathcal{T}_2(x) ) \odot (1-M) \|^2.
\label{eq:r-mae}
\end{equation}
Eq.~\ref{eq:r-mae} jointly optimizes the distance term and the constraint term in Eq.~\ref{eq:disc_mae}. $\lambda$ controls the balance of the two terms. In practice, we let $\phi = \phi'$ to save computational cost, as we empirically find the optimization targets of $d_\phi(\cdot)$ and $d_{\phi'}(\cdot)$ in Eq.~\ref{eq:r-mae} do not diverge very much. 

\paragraph{Empirical evaluation. } First, we verify our claim that MIM representation is robust to image occlusion, as suggested by Eq.~\ref{eq:disc_mae}. We compute the CKA similarity \cite{cka} between the learned features from full images and images with different mask ratios respectively, at each block in the encoder. 
Figure~\ref{fig:cka} shows the CKA similarities of different models. The numbers (0.1 to 0.9) indicate the mask ratios (i.e. percentages of image patches to be dropped) of the test images respectively. 
As shown in Figure~\ref{fig:cka}, both original \emph{MAE} and our relaxed \emph{R-MAE} (as well as another variant \emph{C-MAE}, see the next section) obtain high CKA scores, suggesting those methods learn occlusion invariant features. In contrast, other methods such as supervised training or \emph{MoCo v3} \cite{mocov3} do not share the property, especially if the drop ratio is large. 
After finetuning, the CKA similarities drop, but are still larger than those training from scratch. 

\begin{table}
  \caption{Comparisons of self-supervised methods on ImageNet with ViT-B \cite{dosovitskiy2020image}. \emph{Epochs} in the table indicate numbers of pretraining epochs (for random initialization baselines they are total epochs of training from scratch).
  \emph{PSNR} means the similarity between the generated image (from the masked image) and the original image after pretraining.  } 
  \label{tab:relaxedmae}
  \centering
  \begin{tabular}{l|c|c|c|c|c}
    \toprule
    Pretrain Methods & Transformation & Framework & Epochs & FT Acc (\%) & PSNR (dB) \\
    \midrule
    Random Init & -- & -- & 100 & 80.9 & -- \\ 
     & -- & -- & 300 & 82.1 & -- \\ 
    \midrule
    MoCov3 \cite{mocov3} & crop \& jitter & siamese & 300 & 83.2 & -- \\ % from mae
    DINO \cite{dino} & crop \& jitter & siamese & 800 & 82.8 & -- \\ % from paper
    \midrule
    BeiT \cite{beit} & patch masking & reconstructive & 300 & 82.9 & -- \\
    % MSN \cite{msn} & patch masking & generative & 600 & 83.4 \\
    %MAE \cite{mae} & 300 & 82.9 \\ % from CAE
    MAE \cite{mae} & patch masking & reconstructive & 1600 & 83.6 & 19.3 \\
    CAE \cite{cae} & patch masking & reconst. + siam. & 300 & 83.3 & -- \\ 
    \midrule\midrule
    MAE (\emph{our impl.}) & patch masking & reconstructive & 100 & 83.1 & 22.2 \\
    R-MAE (\emph{ours}) & patch masking & siamese & 100 & 82.7 & 23.7 \\
    \bottomrule
  \end{tabular}
\end{table}

Next, we verify how well \emph{R-MAE} (Eq.~\ref{eq:r-mae}) approximates the original MAE. 
We pretrain the original MAE and R-MAE on ImageNet using the same settings: the mask ratio is 0.75 and training epoch is 100 ($\lambda$ is set to 1 for ours). Then we finetune the models on labeled ImageNet data for another 100 epochs. Results are shown in Table~\ref{tab:relaxedmae}. Our finetuning accuracy is slightly lower than MAE by 0.4\%, which may be caused by the relaxation. Nevertheless, R-MAE roughly maintains the benefit of MAE, which is still much better than supervised training from scratch and competitive among other self-supervised methods with longer pretraining. Another interesting observation is that, the reconstruction quality of R-MAE is even better than the original MAE (see \emph{PSNR} column in Table~\ref{tab:relaxedmae}), which we think may imply the trade-off by the choice of $\lambda$ in Eq.~\ref{eq:r-mae}. We will investigate the topic in the future.

\section{Similarity measurement in MIM is replaceable}

Eq.~\ref{eq:disc_mae} bridges \emph{MIM} and \emph{contrastive learning} with a unified siamese framework. Compared with conventional contrastive learning methods (e.g. \cite{moco,simclr,dino,mocov3,byol}), in MIM two things are special: 1) \emph{data transformations} $\mathcal{T}(\cdot)$: previous contrastive learning methods usually employ random crop or other image jittering, while MIM methods adopt \emph{patch masking}; 2) \emph{similarity measurement} $\mathcal{M}(\cdot,\cdot)$, contrastive learning often uses \emph{InfoNCE} or other losses, while MIM implies a relatively complex\footnote{Notice that the constraint term in Eq.~\ref{eq:disc_mae} also belongs to the similarity measurement.} formulation as Eq.~\ref{eq:new_measurement}. To understand whether the two differences are important, in this section we study how the choice of $\mathcal{M}(\cdot,\cdot)$ affects the performance.

\paragraph{Contrastive MAE (C-MAE).} We aim to replace the measurement $\overline{\mathcal{M}_{\phi, \phi'}}(\cdot, \cdot)$ with a much simpler \emph{InfoNCE loss} \cite{cpc}. We name the new method \emph{contrastive MAE (C-MAE)}. Inspired by \cite{mocov3,byol}, we transform the representations with \emph{asymmetric MLPs} before applying the loss. The new distance measurement is defined as follows:
\begin{equation}
    \widetilde{ \mathcal{M}_{\phi, \phi'}}(z_1, z_2) \triangleq L_{\text{NCE}} = -{\rm log} \frac{{\rm exp}(s(z_1, z_2)/\tau)}{\sum_j{\rm exp}(s(z_1, z'_j)/\tau)}, 
\label{eq:c-mae-measure}
\end{equation}
and 
\begin{equation}
s(z, z') = \frac{q_{\phi'}(p_\phi(z))\cdot p_{\phi}(z')}{\|q_{\phi'}(p_{\phi}(z))\|\cdot\|p_{\phi}(z')\|},
\end{equation}
where $p_\phi(\cdot)$ and $q_{\phi'}(\cdot)$ are \emph{project head} and \emph{predict head} respectively following the name in \emph{BYOL} \cite{byol}, which are implemented with \emph{MLPs}; $\tau$ is the temperature of the softmax. Readers can refer to \cite{mocov3} for details. Hence the objective function of C-MAE is:
\begin{equation}
    L(x, M; \theta, \phi, \phi') = \widetilde{\mathcal{M}_{\phi, \phi'}}(f_\theta(\mathcal{T}_1(x)), f_\theta(\mathcal{T}_2(x))). 
\label{eq:c-mae}
\end{equation}
Unlike Eq.~\ref{eq:disc_mae}, C-MAE does not include nested optimization, thus can be directly optimized without relaxing. 

\paragraph{The design of transformation $\mathcal{T}$.} We intend to use the same transformation as we used in \emph{MAE} and \emph{R-MAE} (Eq.~\ref{eq:new_transformations}). However, we find directly using Eq.~\ref{eq:new_transformations} in C-MAE leads to convergence problem. We conjecture that even though the two transformations derive different patches from the same image, they may share the same color distribution, which may lead to information leakage. Inspired by \emph{SimCLR} \cite{simclr}, we introduce additional color augmentation after the transformation to cancel out the leakage. The detailed color jittering strategy follows \emph{SimSiam} \cite{simsiam}.

\paragraph{Token-wise vs. instance-wise loss. } We mainly evaluate our method on \emph{ViT-B} \cite{dosovitskiy2020image} model. By default, the model generates a latent representation composed of 14x14 patch tokens and one class token, where each patch relates to one image patch while the class token relates to the whole instance. It is worth discussing how the loss in Eq.~\ref{eq:c-mae} applies to the tokens. We come up with four alternatives: apply the loss in Eq.~\ref{eq:c-mae} 1) only to the class token; 2) on the average of all patch tokens; 3) to each patch token respectively; 4) to each patch token as well as the class token respectively. If multiple tokens are assigned to the loss, we gather all loss terms by averaging them up. Table~\ref{tab:ablation} shows the ablation study results. It is clear that token-wise loss on the patch tokens achieves the best finetuning accuracy on \emph{ImageNet}. In comparison, adding the class token does not lead to improvement, which may imply that class token in self-supervised learning is not as semantic as in supervised learning. Therefore, we use a token-wise-only strategy for C-MAE by default.

\paragraph{Implementation details.} Following \cite{mocov3}, we use a siamese network, which contains an online model and a target model whose parameters are EMA updated by the online model. We use 2-layer projector (i.e. $p_\phi(\cdot)$ in Eq.~\ref{eq:c-mae}) and 2-layer predictor ($q_{\phi'}$), and use GELU as activation layer. 
To represent the masked patches into the encoder network, we adopt \emph{learnable mask tokens} as \cite{simmim,beit} does rather than directly discard the tokens within the masked region as the original MAE, because unlike MAE, our C-MAE does not include a heavy transformer-based decoder to predict the embeddings for the masked region.

\paragraph{Result and discussion.} Table~\ref{tab:tokenwisemoco} shows the finetuning results of C-MAE and a few other self-supervised methods. C-MAE achieves comparable results with the counterpart MAE baselines, suggesting that \emph{in MIM framework the reconstructive decoder, or equivalently the measurement in siamese form (Eq.~\ref{eq:new_measurement}),  does not matter much}. A simple \emph{InfoNCE loss} works fine. We also notice that our findings agree with recent advances in \emph{siamese MIMs}, \emph{e.g.} \emph{iBOT} \cite{zhou2021ibot}, \emph{MSN} \cite{msn} and \emph{data2vec} \cite{baevski2022data2vec}, whose frameworks involve various distance measurements between the siamese branches instead of reconstructing the unmasked parts, however, achieve comparable or even better results than the original reconstruction-based MIMs like \cite{mae,beit}. In addition to those empirical observations, our work uncovers the underlying reason: both reconstructive and siamese methods target learning occlusion invariant features, thereby it is reasonable to obtain similar performances.

Table~\ref{tab:tokenwisemoco} also indicates that, as siamese frameworks, C-MAE achieves comparable or even better results than previous counterparts such as \emph{DINO} \cite{dino}, even though the former mainly adopts random patch masking while the latter involves complex strategies in \emph{data transformation}. \cite{mae} also reports a similar phenomenon that data augmentation is less important in MIM. The observation further supports the viewpoint that \emph{learning occlusion invariant feature is the key to MIM, rather than the loss. } Intuitively, to encourage occlusion invariance, patch masking is a simple but strong approach. For example, compared with random crop strategy, patch masking is more general -- cropping can be viewed as a special mask pattern on the whole image, however, according to the experiments in \cite{simmim,mae}, it is good enough or even better to leave patch masking fully randomized\footnote{Although very recent studies \cite{shi2022adversarial,kakogeorgiou2022hide,li2022semmae,wu2022object} suggest more sophisticated masking strategies can still help.}.

\begin{table}
  \caption{Comparisons of C-MAE and other pretraining methods on ImageNet finetuning. All models are based on ViT-B.}
  \label{tab:tokenwisemoco}
  \centering
  \begin{tabular}{l|c|c}
    \toprule
    Pretrain Methods & Epochs & FT Acc (\%) \\
    \midrule
    Random Init & -- & 80.9 \\ 
    \midrule
    MoCo v3 \cite{mocov3} & 300 & 83.2 \\ % from mae
    DINO \cite{dino} & 800 & 82.8 \\ % from paper
    MAE \cite{mae} & 1600 & 83.6 \\
    \midrule
    \midrule
    MAE (\emph{our impl.}) & 100 & 83.1 \\
    C-MAE (\emph{ours}) & 100 & 82.9 \\
    MAE (\emph{our impl.}) & 400 & 83.2 \\
    C-MAE (\emph{ours}) & 400 & 83.1 \\
    \bottomrule
  \end{tabular}
\end{table}

\begin{table}
  \caption{Ablation study on the strategies of C-MAE loss. All models are pretrained for 100 epochs.}
  \label{tab:ablation}
  \centering
  \begin{tabular}{l|c|c|c}
    \toprule
    Measurement & Class Token & Patch Tokens & FT Acc (\%) \\
    \midrule
    instance-wise & \checkmark & & 82.5 \\
    instance-wise & & average & 82.6 \\
    token-wise  & & \checkmark & 82.9 \\
    token-wise & \checkmark & \checkmark & 82.9 \\
    \bottomrule
  \end{tabular}
\end{table}

\paragraph{Additional ablations.} Table~\ref{tab:addi_ablation} presents additional results on \emph{MAE} and \emph{C-MAE}. First, Although C-MAE shows comparable fine-tuning results with MAE, we find under \emph{linear probing} \cite{moco,mae} and \emph{few-shot} (i.e. fine-tuning on 10\% ImageNet training data) protocols, C-MAE models lead to inferior results. Further study shows the degradation is mainly caused by the usage of mask tokens in C-MAE, which is absent in the original MAE -- if we remove the mask tokens as done in MAE's encoder, linear probing and few-shot accuracy largely recover (however fine-tuning accuracy slightly drops), which we think is because mask tokens enlarge the structural gap between pretraining and linear/few-shot probing, since the network is not fully fine-tuned under those settings. 

Second, we further try replacing the \emph{InfoNCE} loss \ref{eq:c-mae-measure} with \emph{BYOL} \cite{byol} loss in \emph{C-MAE}. Following the ablations in Table~\ref{tab:ablation}, we still make the BYOL loss in \emph{token-wise} manner. Compared with InfoNCE, BYOL loss does not have explicit negative pairs. Results imply that BYOL loss shows similar trend as InfoNCE loss, which supports our viewpoint ``similarity measurement in MIM is replaceable''. However, we also find BYOL loss is less stable, resulting in slightly lower accuracy than that of InfoNCE.

Last, since our \emph{C-MAE} involves \emph{color jittering} \cite{simclr}, one may argue that color transformation invariance could be another key factor other than occlusion invariance. Unfortunately, the ablation study is nontrivial because we find the contrastive loss quickly collapses without color jittering. So instead, we study the original \emph{MAE} with additional color jittering. We compare two configurations: a) augmenting the whole image before applying MAE; b) only augmenting the unmasked patches (i.e. the reconstruction targets keep the same). Results show that neither setting boosts MAE further, which implies the invariance of color jittering does not matter much. 

\begin{table}[t]
  \caption{Additional comparisons on MAE and C-MAE. All models are pretrained and fine-tuned for 100 epochs respectively.}
  \label{tab:addi_ablation}
  \centering
  \begin{tabular}{l|c|c|c}
    \toprule
    Pretrain Methods & Lin. Prob Acc (\%) & FT Acc (\%) & 10\% FT Acc (\%) \\
    \midrule
    MAE & 54.5 & 83.1 & 67.5 \\
    MAE w/ color jitter (whole image) & 53.4 & 83.1 & 67.3 \\
    MAE w/ color jitter (unmasked only) & 54.0 & 83.0 & 67.3 \\
    \midrule
    C-MAE & 41.1 & 82.9 & 66.4 \\
    C-MAE w/o mask token & 56.2 & 82.6 & 67.5 \\
    \midrule
    C-MAE (BYOL loss) & 26.9 & 82.8 & 65.2 \\
    C-MAE w/o mask token (BYOL loss) & 55.2 & 82.5 & 66.1 \\
    \bottomrule
  \end{tabular}
\end{table}

\section{MIM can learn a favored, (almost) data-agnostic initialization}
\label{sec:semantic_info} 

As discussed in the above sections, learning occlusion invariant features is the key ``philosophy'' of \emph{MIM} methods. Hence an interesting question comes up: how do the learned networks model the invariance? One possible hypothesis is that occlusion invariance is represented in an \emph{data-agnostic} way, just analogous to the structure of \emph{max pooling} -- the output feature is robust only if the most significant input part is not masked out, thereby the invariance is obtained by design rather than data. Another reasonable hypothesis is, in contrast, the invariance requires knowledge from a lot of data. In this section we investigate the question. 

Inspired by \cite{train_one_image}, to verify our hypotheses we try to \emph{significantly} reduce the number of images for \emph{MAE} pretraining, i.e. ranging from 1 for 1000 randomly sampled from \emph{ImageNet} training set, hence the semantic information from training data should be very limited in the pretraining phase. Notice that MAE training tends to suffer from over-fitting on very small training set, as the network may easily ``remember'' the training images. Therefore, we adopt stronger data augmentation and early-stop trick to avoid over-fitting. Table~\ref{tab:strong_aug} presents the result. Very surprisingly, we find pretraining with only \emph{one} image with 5 epochs already leads to improved finetuning score -- much better than 100-epoch training from scratch and on par with training for 300 epochs. The fine-tuning results do not improve when the number of pretrain images increases to 1000. Since it is not likely for only one image to contain much of the semantic information of the whole dataset, the experiment provides strong evidence that MIM can learn a favored initialization, more importantly, which is (almost) data-agnostic. Table~\ref{tab:sampling} also indicates the choice of sampling strategy does not affect the fine-tuning accuracy, further suggesting that such benefit from MIM pre-training might be free of category information. 

\begin{table}
  \caption{Comparisons of MAE pretrained with different numbers of images.} 
  \label{tab:strong_aug} 
  \centering 
  \begin{tabular}{l|c|c|c|c} 
    \toprule 
    Pretrain Images & Stronger Aug. & Train Epochs & FT Epochs & FT Acc (\%) \\
    \midrule
    1 & \textbf{\checkmark} & 5 & 100  &  \textbf{82.3} \\ 
    \multirow{2}{*}{10} &  & 2 & 100 & 81.9 \\ 
     & \checkmark & 5 & 100 & 82.3 \\ 
    \multirow{2}{*}{100} &  & 10 & 100 & 82.1 \\ 
     & \checkmark & 10 & 100 & 82.2 \\ 
    \multirow{2}{*}{1000} &  & 100 & 100 & 82.2 \\ 
     & \checkmark & 100 & 100 & 82.2 \\ 
     \midrule
    Random Init & & - & 100 & 80.9 \\
     & & - & 300 & 82.1 \\
    \bottomrule
  \end{tabular} 
\end{table}

\begin{table}[t]
  \caption{Comparisons of different image sampling strategies. For MAE pretraining, 1000 images are sampled with different strategies respectively from ImageNet.}
  \label{tab:sampling}
  \centering
  \begin{tabular}{l|c|c}
    \toprule
    Sampling Strategy & \# of Categories & FT Acc (\%) \\
    \midrule
     in one class & 1 & 82.2 \\
     random & 617 & 82.2 \\
     one per class & 1000 & 82.2 \\
    \bottomrule
  \end{tabular}
\end{table}

\begin{table}[t]
  \caption{Comparisons of different pretraining methods on 1000 images sampled from ImageNet (one image for each class). All methods pretrain for 100 epochs on the sampled dataset (except for random initialized baseline) and then fine-tune for 100 epochs on full/10\%-ImageNet accordingly. }
  \label{tab:few_image}
  \centering
  \begin{tabular}{l|c|c|c}
    \toprule
    Pretrain Methods & Lin. Prob Acc (\%) & FT Acc (\%) & 10\% FT Acc (\%) \\
    \midrule
    Random Init & 6.1 & 80.9 & 34.9 \\
    \midrule
    Supervised & 33.1 & 81.0 & 52.6 \\
    MoCo v3 & 37.3 & 79.2 & 45.8 \\
    \midrule
    MAE &  13.8 & 82.2 & 57.6 \\
    R-MAE & 25.9 & 82.1 & 58.8 \\
    C-MAE & 20.1 & 82.1 & 61.9 \\
    \bottomrule
  \end{tabular}
\end{table}

Moreover, in Table~\ref{tab:few_image} we benchmark various pretraining methods on a 1000-image subset from \emph{ImageNet} training data, which provides more insights on MIM training. We find the linear probing accuracy of \emph{MAE} is very low, which is only slightly better than random feature (first row), suggesting that the feature learned from 1000 images is less semantic; however, the finetuning result as well as few-shot fine-tuning is fine. Our proposed \emph{R-MAE} and \emph{C-MAE} share similar properties as the original MAE -- relatively low linear probing scores but high fine-tuning performance. The observation strongly supports our first hypothesis at the beginning of Sec.~\ref{sec:semantic_info}: the occlusion invariance learned by MIM could be data-agnostic, which also serves as a good initialization for the network. In comparison, supervised training and \emph{MoCo v3} \cite{mocov3} on 1000 images fail to obtain high fine-tuning scores, even though their linear probing accuracy is higher, which may be because those methods cannot learn occlusion-invariant features from small dataset effectively. In the appendix, we will discuss more on the topic. 

\section{Experimental Details}

\paragraph{Pretraining.}
We use ViT-B/16 as the default backbone. For MAE pretraining, we use the same settings as \cite{mae}, and use the patch normalization when computing loss. We use the mask ratio of 0.75, which is the most effective one in \cite{mae}. We use AdamW optimizer with cosine decay scheduler and the batch size is set to 1024. We set the base learning rate (learning rate for batch size of 256) as 1.5e-4 with a 20-epoch linear warm-up and scale up the learning rate linearly when batch size increases \cite{linear_rule}. 
For R-MAE, we search the learning rate and finally set the base learning rate as 3.0e-4. Other training settings are the same as \cite{mae}. 
For C-MAE, the momentum to update the teacher model is set to 0.996, and the temperature to compute contrastive loss is set to 0.2. For projector and predictor heads, we set 2048-d for hidden layers. We search the learning rate and finally set the base learning rate as 1.5e-4. Other parameters are the same as C-MAE. 
We train the model for 100 epochs on the ImageNet~\cite{imagenet} dataset as default. 
Due to the computational resource constraints, we report the results of 400 epochs to prove that our method gains better results with longer training.

\paragraph{Finetuning.}
We follow the training settings in \cite{mae}. We use the average pooling feature of the encoded patch tokens as the input of classifier, and train the model end-to-end. Following\cite{mae}, we reset the parameters of the final normalization layer. We use AdamW optimizer with cosine decay scheduler and set the batch size to 1024. We set the base learning rate as 1.0e-3 with 5-epoch linearly warm-up and train the model for 100 epochs. 
Note that the supervised trained ViT in our paper uses the same settings as finetuning and the model is trained for 100 epochs.

\section{Related Work}

\paragraph*{Masked Image Modeling.} 
As the ViT models achieve breakthrough results in computer vision, self-supervised pretraining for ViTs becomes an intense scholarly domain. In addition to siamese frameworks such as \cite{mocov3,dino}, MIM is an efficient and popular way of self-supervised modeling. 
The model learns rich hidden information by optimizing the reconstruction model \cite{mae}. Following BERT \cite{bert}, \cite{igpt} compress the image to a few pixels, and then directly learn the masked pixel color. \cite{beit} maps all image patches to 8192 embeddings by training d-VAE \cite{dall_e}, and then learns the correct embedding correspondence for mask patches.
\cite{mst} optimizes the masking process based on BEiT. \cite{el2021large,zhou2021ibot,mst} combines MIM with siamese frameworks and improves the performance of linear probing. \cite{mae, simmim} use a simple method to reconstruct the original image, and also learn rich features effectively. \cite{understand_mae} gives a mathematical understanding of MAE. MSN \cite{msn}, which is a concurrent work of ours, also discusses the invariance to mask. 

\paragraph*{Siamese approaches in SSL.}
Self-supervised pretraining achieves great success in classification \cite{exemplar_cnn,context_prediction,rotation,jigsaw,cpc,inst_disc,moco,simclr,byol,barlowtwins}, detection\cite{self_emd,xiong2020loco,lang2021contrastive,xie2021detco} and segmentation. One of the promising methods is based on siamese frameworks \cite{cmc,moco,mocov2,mocov3,pirl,simclr,mocov2,byol,swav,simsiam,propagate_yourself,dino,barlowtwins,vicreg}, which learns representations by minimizing the distance of positive samples with siamese networks. 
In practice, \cite{simclr, swav, simsiam, barlowtwins} uses the same parameters in the online and target model, while \cite{moco, mocov2, mocov3, byol, dino} updates online parameters to target using exponential moving average. Only minimizing the distance of positive samples will cause the model to fall into trivial solutions, so a critical problem in SSL is how to prevent such a model from collapsing. \cite{simclr, moco} use negative samples from different images, then computes contrastive loss. \cite{byol, simsiam} add an extra predictor on the top of the online model then stop the gradient of the target model. Instead of optimizing the loss per instance, \cite{barlowtwins, vicreg} optimize the variance, covariance or cross-covariance on the channel dimension. \cite{dino} optimize the distributions of the two features, and avoid trivial solutions by centering and sharpening. 

\section{Limitation}
When implementing RelaxMIM, we only used MAE as the backbone and do not try other MIM methods. When implementing discriminative MIM, we simply use InfoNCE, which we believe can be replaced by other contrastive learning methods. Due to the lack of computational resources, we only use ViT-B as the backbone and train all models on ImageNet-1k, and our models are all trained for much fewer epochs than commonly used in self-supervised methods. In our future work, we plan to train the models longer and use larger scale models (such as ViT-L) to get better results.

\section{Conclusion}

In this paper, we propose a new viewpoint: MIM implicitly learns occlusion-invariant features, and build up a unified understanding framework \emph{RelaxMIM} for MIM and contrastive learning. In the view of RelaxMIM, MIM models intrinsically learn \emph{occlusion invariant features}. Then we verify that the representation of RelaxMIM is robust to image occlusion. Based on RelaxMIM, we replace the similarity measurement with simpler InfoNCE loss and achieve comparable results with the original MIM framework. It suggests that \emph{patch masking} may be the critical component of the framework. To understand why patch masking is important, we perform MIM pretraining on very few images and finetune the encoder with supervised training on full ImageNet. We find that the encoder learns almost \emph{data-agnostic} occlusion invariant features during pretraining, which could be a favored initialization for finetuning. To measure whether the MIM method has learned human recognition patterns, we compare the shape bias of different self-supervised models and conclude that,  MIM could improve the recognition ability of ViT to make it closer to human recognition, but the improvement may be limited. We hope the RelaxMIM framework may inspire more powerful self-supervised methods in the community.

\small

% Supplementary
\newpage
\appendix

\section{More Visualization Experiments}

\subsection{Occlusion-invariance of Few Images Pretrained MAE}
Here we discuss the occlusion invariance of a few images pretrained MAE models. We use \textbf{CKA similarities} between the representations generated by the masked image and the full image under different mask ratios as protocol.  
The numbers (0.1 to 0.9) indicate the mask ratios (i.e. percentages of image patches to be dropped) of the test images respectively.
The higher CKA similarity with a large mask ratio means the model learns better occlusion invariance.

\begin{figure}[H]
  \centering
  \includegraphics[width=1.0\textwidth]{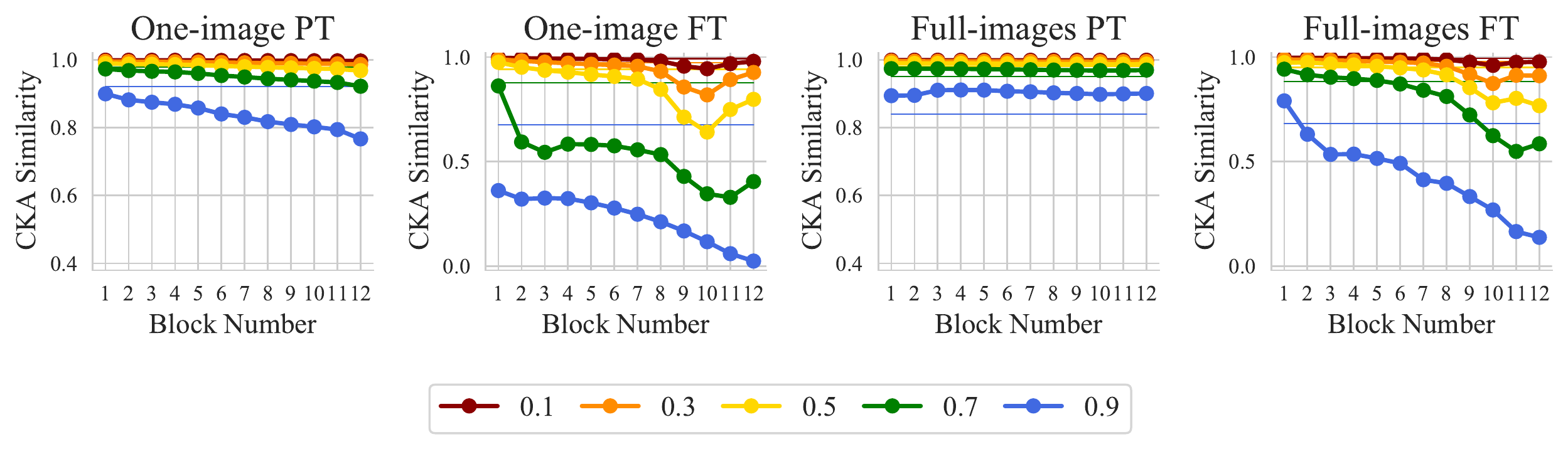}
  \caption{CKA similarity between the representations generated by the masked image and the full image respectively under different mask ratios. \textbf{One-image PT} indicates the model that pretrained on one image for 5 epochs, and \textbf{Full-images PT} indicates the model pretrained on ImageNet training set for 100 epochs. The finetuning models (\textbf{FT}) are all trained on ImageNet training set for 100 epochs.}
  \label{fig:cka_single_image}
\end{figure}

Figure \ref{fig:cka_single_image} shows the CKA similarities of MAE pretrained with different amounts of data.  As the figures show, the model learns occlusion invariance even pretrained with one image. Unfortunately, the model does not keep the occlusion invariance after finetuning. When the mask ratio increase to 0.7, the CKA similarities drop significantly below 0.5. In Comparison, full-set pretrained MAE is not so sensitive to the change of mask ratio (after 0.7) after finetuning.  

\begin{figure}
  \centering
  \includegraphics[width=1.0\textwidth]{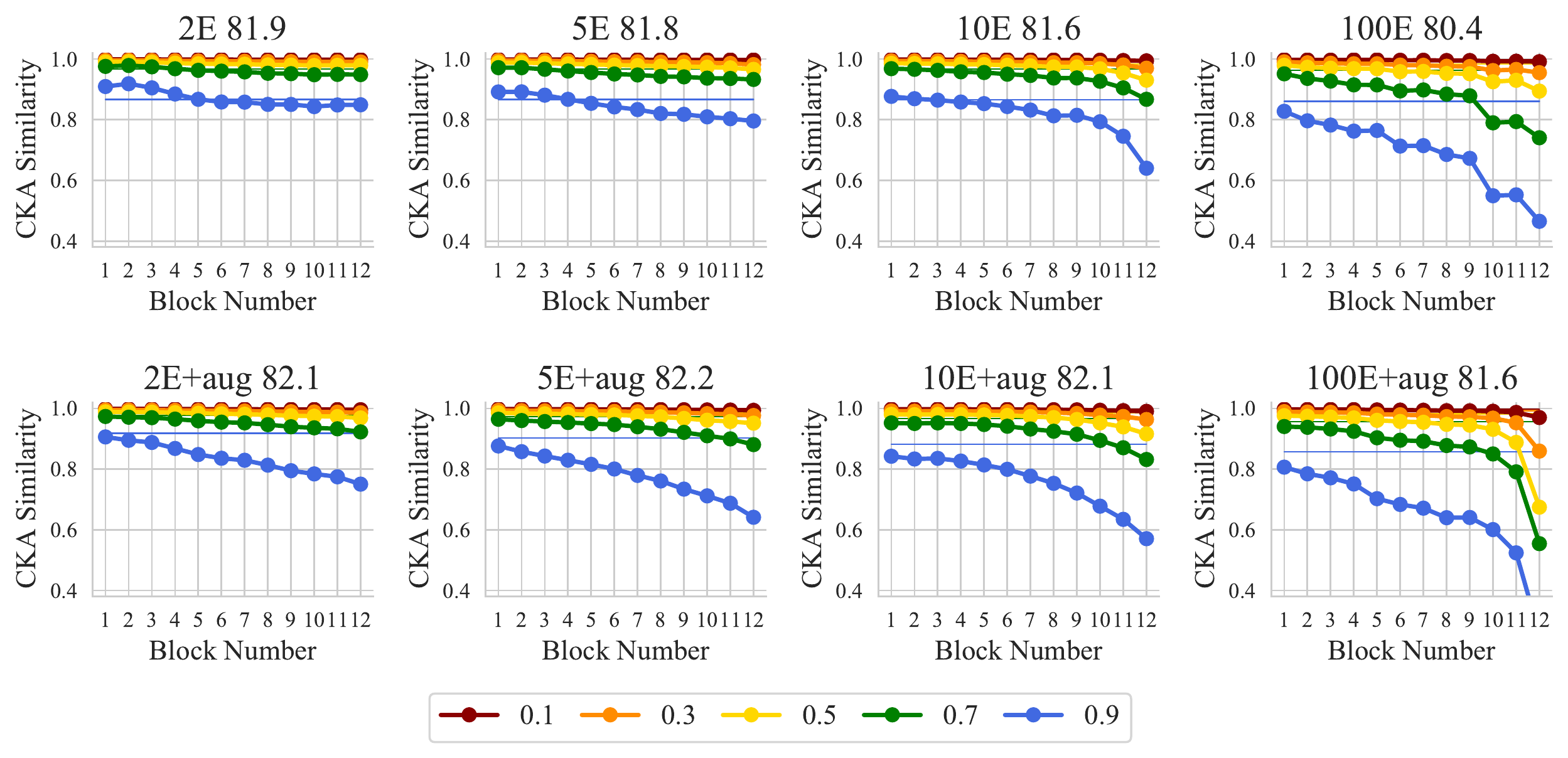}
  \caption{CKA similarity between the representations generated by the masked image and the full image respectively under different mask ratios. The number on the subtitle is the finetuning result of the model. All models are pretrained on 10 images for different epochs and finetuned on full ImageNet training set for 100 epochs. \textbf{$N$E} means the model pretrained for $N$ epochs. \textbf{+aug} means adding stronger augmentations.}
  \label{fig:cka_10img}
\end{figure}

Furthermore, we discuss the relationship between occlusion invariance with overfitting.
We train the MAE with different training epochs on 10 images and plot the CKA similarities. Results in Figure~\ref{fig:cka_10img} show that, overfitting affects the learning of occlusion invariance, and causes the performance to drop. We further explore the way to prevent overfitting, \textit{using stronger data augmentation}, whether beneficial to maintain occlusion invariance. As shown in Figure~\ref{fig:cka_10img}, even the finetuning results increase a little when using stronger augmentations, the occlusion invariance does not been improved. 

\subsection{Comparison with Human Recognition}
\begin{figure}
  \centering
  \includegraphics[width=0.8\textwidth]{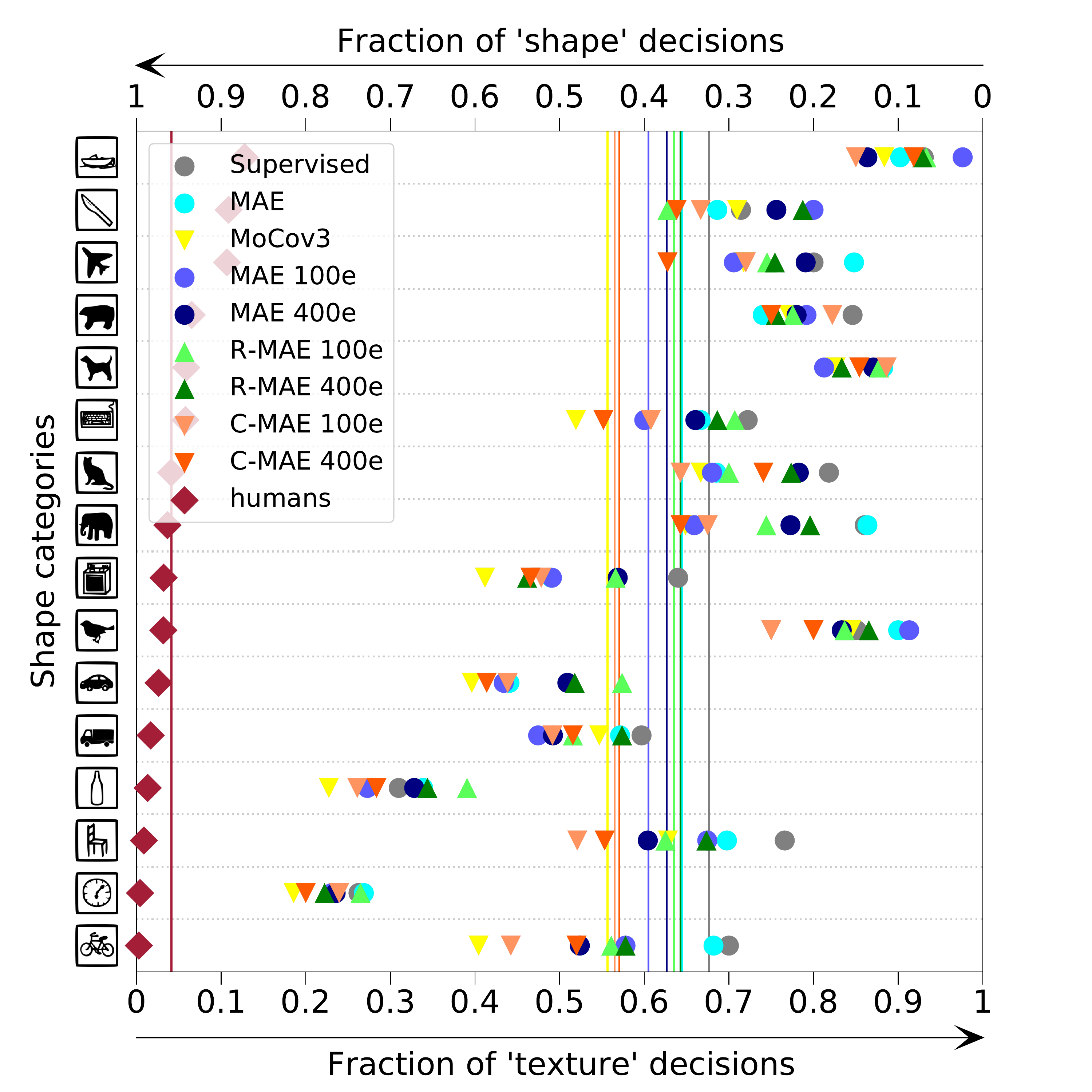}
  \caption{Shape bias of MAE, MoCov3, R-MAE and C-MAE pretrained on ImageNet. The vertical line is the average shape bias of 16 classes.}
  \label{fig:shape_bias}
\end{figure}

\cite{2105_07197} shows that, ViT behaves more like humans in classification, and we wonder whether our proposed siamese framework learns more high-level perception. 
Following the method in \cite{2105_07197}, we plot the shape bias of MIM models in Figure~\ref{fig:shape_bias}.

Figure~\ref{fig:shape_bias} shows the shape bias of MAE, MoCov3, R-MAE and C-MAE. 
As shown in the figure, the grey line represents the supervised trained model, which has the lowest shape bias. That means fully supervised learning prefers to learn texture information rather than self-supervised pretrained models.
Both MAE (blue line) and R-MAE (green line) learn less shape bias than MoCo (yellow line) and C-MAE (orange line). We speculate that it is because the target of the pretext task of MIM is closer to the original images (or exactly the origin images), which makes the model learn more texture features.
Additionally, C-MAE learns a similar shape-bias compared with MoCo~v3. The results indicate that instance-wise learning is not necessary for models to learn as human does,  learning occlusion invariance could also improve the ability of the model to learn shape-bias.
When training longer, all masked-based models are biased to learn texture features. We conclude that the masked-based models could learn the ability to complete object shape quickly in a few epochs, and then learn to reconstruct the texture of images. 
\clearpage

%%%%%%%%%%%%%%%%%%%%%%%%%%%%%%%%%%%%%%%%%%%%%%%%%%%%%%%%%%%%

\end{document}